\documentclass[conference]{IEEEtran}
\IEEEoverridecommandlockouts
\usepackage{cite}
\usepackage{amsmath,amssymb,amsfonts}
\usepackage{graphicx}
\usepackage{textcomp}
\usepackage{float}
\usepackage{hyperref}
\usepackage{multirow}
\usepackage{algorithm}
\usepackage[noend]{algpseudocode}
\usepackage{stfloats}
\usepackage{subfigure}
\usepackage{caption}
\usepackage{setspace}
\usepackage{hyperref}
\usepackage{cuted}
 
\usepackage{lipsum}

\usepackage{xcolor}
\def\BibTeX{{\rm B\kern-.05em{\sc i\kern-.025em b}\kern-.08em
    T\kern-.1667em\lower.7ex\hbox{E}\kern-.125emX}}
\begin{document}

\title{
ClickAttention: Click Region Similarity Guided Interactive Segmentation  \\
\thanks{Identify applicable funding agency here. If none, delete this.}
}

\author{\IEEEauthorblockN{Long Xu}
\and
\IEEEauthorblockN{Shanghong Li}
\and
\IEEEauthorblockN{Yongquan Chen}
\and
\IEEEauthorblockN{Junkang Chen}
\and
\IEEEauthorblockN{Rui Huang}
\and
\IEEEauthorblockN{Feng Wu}
}

\twocolumn[{
\renewcommand\twocolumn[1][]{#1}
\maketitle
\begin{center}
    \captionsetup{type=figure}
    \includegraphics[scale=0.3]{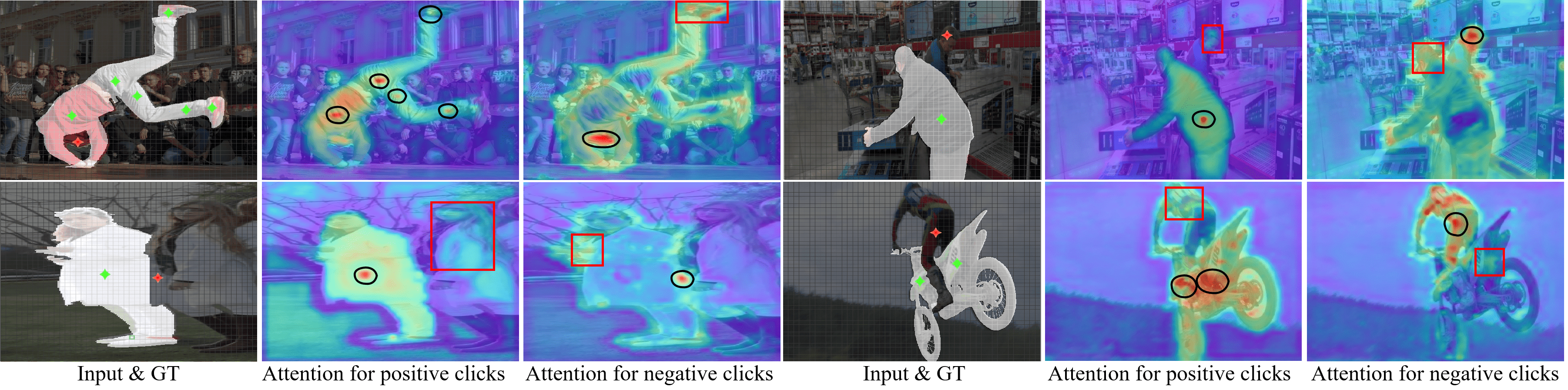}
    \captionof{figure}{Attention visualization of clicked regions based on Transformer \cite{dosovitskiy2020image, 9878519, liu2022simpleclick}. \textcolor{red}{Red marks} denote the negative click; \textcolor{green}{Green marks} indicate the positive click; \textbf{Black} circles indicate the limited click influence range issue; \textcolor{red}{Red rectangles} highlight the regions that should not be activated in the attention maps of positively/negatively-clicked regions. }
    \label{fig:teaser}
\end{center}
}]

\begin{abstract}
Interactive segmentation algorithms based on click points have garnered significant attention from researchers in recent years.
However, existing studies typically use sparse click maps as model inputs to segment specific target objects, which primarily affect local regions and have limited abilities to focus on the whole target object, leading to increased times of clicks.
In addition, most existing algorithms can not balance well between high performance and efficiency.
To address this issue, we propose a click attention algorithm that expands the influence range of positive clicks based on the similarity between positively-clicked regions and the whole input. 
We also propose a discriminative affinity loss to reduce the attention coupling between positive and negative click regions to avoid an accuracy decrease caused by mutual interference between positive and negative clicks.
Extensive experiments demonstrate that our approach is superior to existing methods and achieves cutting-edge performance in fewer parameters. 
An interactive demo and all reproducible codes will be released at \url{https://github.com/hahamyt/ClickAttention}

\end{abstract}


\section{Introduction}
Interactive segmentation is a critical task in human-computer interaction and data annotation that enables target object segmentation in an image with limited user interaction.
This paper focuses on click-based interactive segmentation to achieve the target accuracy with minimal parameters and clicks.

Current click-based methods typically represent user clicks via a click map that is concatenated with the input image for target object segmentation \cite{7982754,7108034,7780416,9187581,7544505,9157109,DBLP:journals/corr/abs-2003-07932,9156403,9607681,9709986,9878519,liu2022pseudoclick,liu2022simpleclick}.
\cite{9157109,DBLP:journals/corr/abs-2003-07932,9709986} adopted Gaussian click maps to model user clicks, 
where the map only had effective values within a small range of fixed variance near the clicked point,
thus only affecting a limited range of the target object.
Majumder Soumajit, et al. \cite{majumder2019content} recognized the limitations of these methods and used the hierarchical structural prior in the image to improve segmentation performance.
Ding Zongyuan, et al. \cite{9729600} utilized segmentation results to generate new click points and further optimized the target mask with these click points.
Recently, Kirillov, et al. \cite{kirillov2023segment} proposed a SAM model, which adopted the sparse prompt embedding \cite{tancik2020fourier} to directly encode the coordinates of clicks, which is different from the methods mentioned above.
Although these methods have built better click models to some extent, they still struggle to address the issue of limited click influence range and efficiency.

Since positive and negative clicks indicate the target and the background respectively, Xu Ning, et al. \cite{7780416} constructed a two-channel click map, with one channel designated for positive clicks and the other for negative clicks. 
Though widely adopted in subsequent works \cite{9157109,DBLP:journals/corr/abs-2003-07932,9156403,9607681,9709986,9878519,liu2022pseudoclick,liu2022simpleclick},
this strategy exacerbates the sparsity of clicks and hinders the network's ability to segment accurately.

Recently, SAM \cite{kirillov2023segment} has received widespread attention from people.
Ke Lei, et al. \cite{sam_hq} proposed an HQ-SAM, which brings higher segmentation quality than SAM by using an adapter.
Similarly, Tianrun Chen, et al. \cite{chen2023sam} proposed a SAM-based SAM-adapter model to improve the performance of SAM in underperformed scenes.
However, these models have a large number of parameters, resulting in low efficiency. Also, their input size is 1024x1024, leading to a high number of FLOPS than current works (maximum is 672x672).
To further improve the speed of SAM, Xu Zhao, et al.\cite{zhao2023fast} proposed a FastSAM based on YOLOv8, which reduces the parameters of SAM and GPU memory usage significantly. 
However, these methods still suffer from inefficiency issues due to the large and complex model, which limits its application in low-end devices.

\begin{figure}[!t]
    \centering
    \includegraphics[scale=0.71]{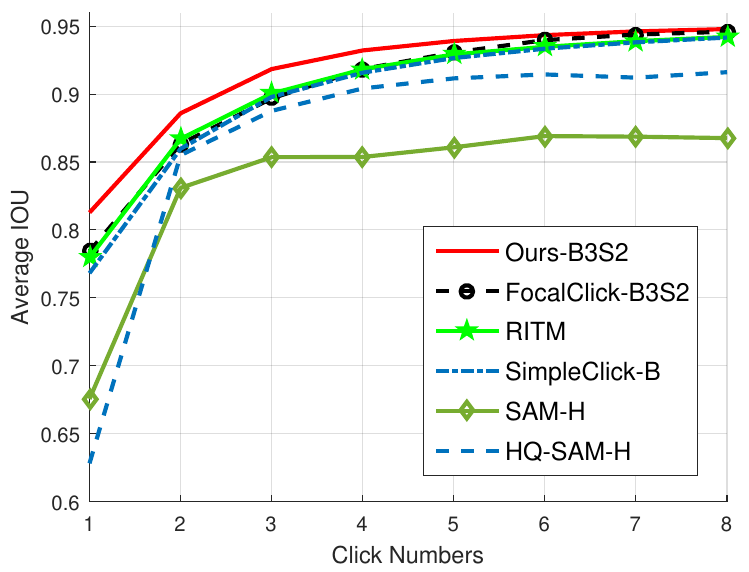}
    \caption{The average IOU varies with clicks (based on the average results of 8 benchmarks), indicating the proposed method can utilize fewer clicks to obtain better precision}
    \label{fig:click_analysis}
\end{figure}

To investigate the role of user clicks in interactive segmentation, we use the explainable visualization method for the self-attention of models, as proposed in \cite{10132428}. 
Specifically, we visualize the attention maps of positively/negatively-clicked regions as shown in Figure \ref{fig:teaser}.
The black circles mark the positively-clicked local regions, indicating the areas where the model allocates more attention while de-emphasizing the rest of the target object.
Notably, the activated attention ranges of positively and negatively clicked regions overlap in multiple areas.
Therefore, we use a red rectangle to highlight the overlapping regions that should not be activated in the attention maps by positive or negative clicks.

To address these issues, we propose an image patch-based \cite{dosovitskiy2020image} click attention algorithm and an affinity loss inspired by SASFormer \cite{su2022sasformer}.
This novel approach aims to decouple positive and negative clicks, guiding positive ones to focus on the target object and negative ones on the background.
To evaluate its effectiveness, we assessed the proposed algorithm on widely used benchmarks, including Grabcut \cite{10.1145/1015706.1015720}, Berkeley \cite{MCGUINNESS2010434}, DAVIS \cite{perazzi2016benchmark}, SBD \cite{hariharan2011semantic},
COCO-MVal \cite{10.1007/978-3-319-10602-1_48},
PascalVOC \cite{everingham2009pascal}, and
DAVIS-585 \cite{9878519}.

In conclusion, this paper presents the following contributions to interactive segmentation:

\begin{itemize}
    \item We propose a click attention algorithm based on local region similarity, which can effectively model the intents behind user clicks and reduce the number of clicks.
    \item We propose a discriminative attention affinity loss, which can effectively decouple positive and negative clicks and improve segmentation accuracy.
    \item As shown in Figure \ref{fig:click_analysis}, the proposed algorithm achieves state-of-the-art (SOTA) performance.
\end{itemize}

\section{Related Work}
\subsection{Interactive segmentation}
Interactive segmentation algorithms increasingly rely on deep learning. 
Traditional methods \cite{1704833,gulshan2010geodesic,kim2010nonparametric,10.1145/1015706.1015720} have been surpassed by deep learning methods \cite{Zhang2022IntentionawareFP,9878519,liu2022pseudoclick,liu2022simpleclick,7780416, 8953578, DBLP:journals/corr/abs-2003-07932, 9897365, 9607681,9729600,Zhang2022IntentionawareFP,liu2022simpleclick} which become the current research mainstream because of their excellent performance.

To mitigate the problem of limited click influence range, existing studies concentrate on modifying the input of the network.
For instance, Sofiiuk Konstantin, et al. \cite{9897365} utilized the network's previous output to simulate user intent, but this strategy suffers from lags and difficulties in modeling individual clicks, which can increase the number of required clicks.
Liu Qin, et al. \cite{liu2022simpleclick} used MAE-ViT \cite{DBLP:journals/corr/abs-2111-06377} to handle multimodal information and improve performance by increasing network parameters. 
However, this model has high computational requirements, which limits its deployment on low-end devices.
Chen Xi, et al. \cite{9878519} expanded the influence range of a single click by further processing the region of interest (ROI). 
However, this approach discards click points in other regions and relies heavily on the accuracy of previous segmentation.
Most works \cite{8953578,Zhang2022IntentionawareFP,liu2022simpleclick, 9878519} have used a two-channel click map to differentiate between positive and negative clicks, which exacerbates sparsity and fails to address the coupling problem between the two types of clicks, resulting in imprecise segmentation.

Although existing works incorporate user clicks in different ways to improve, inherent challenges remain.

\subsection{Attention mechanism in local regions}
In recent years, attention mechanisms have fueled success across multiple visual tasks.

Fu Jun, et al. \cite{8953974} significantly improved the segmentation performance by constructing rich contextual dependencies on local region features.
Schlemper Jo, et al. \cite{SCHLEMPER2019197} implemented an attention gate mechanism to suppress the network's focus on irrelevant regions, which accelerated the learning of task-relevant features.
The design of the attention mechanism in Vision Transformer \cite{dosovitskiy2020image} can capture the correlation between patches, enabling the network to focus on specific areas.
Ma Jie, et al. \cite{10132428} investigated the interpretability of the self-attention mechanism in Vision Transformer and proposed a method to analyze attention interactions between different patches. This method can guide the model to focus on specific areas by adaptively adjusting the network.

These studies inspired us to use attention mechanisms to bridge the gap between user clicks and segmentation results, guiding the network to efficiently understand user intent.

\section{Methods}
\begin{figure*}[!t]
    \centering
    \includegraphics[scale=0.8]{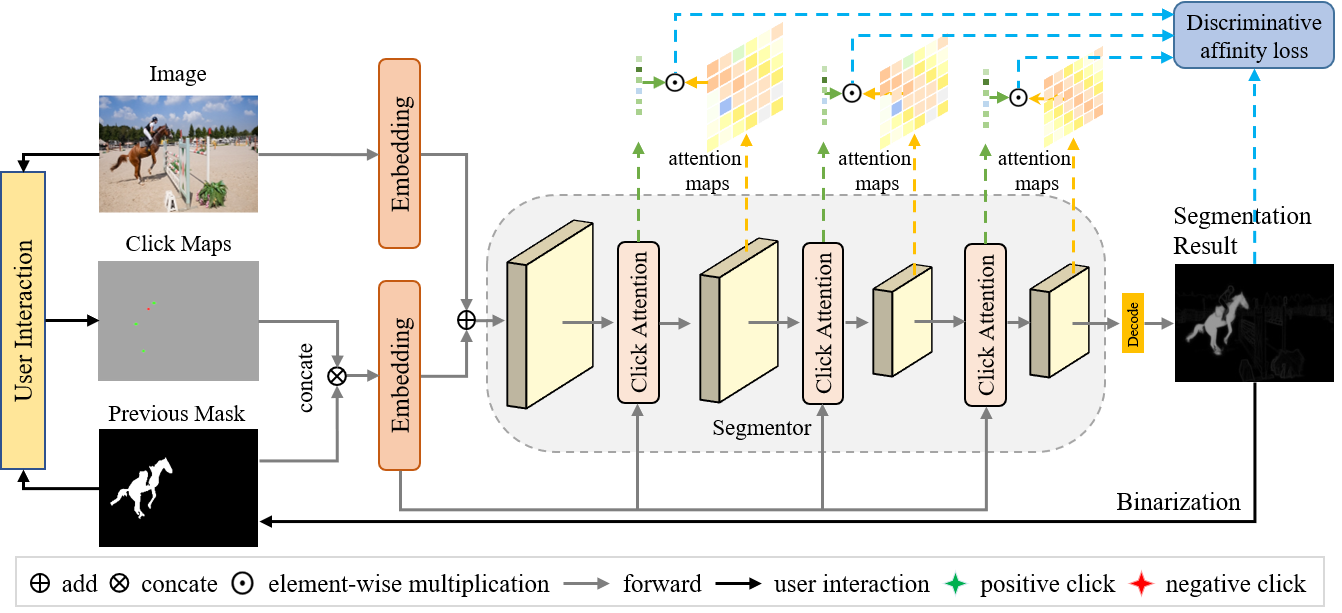}
    \caption{Overall framework of the proposed algorithm. Four downsampling blocks come from Segformer \cite{NEURIPS2021_64f1f27b}. Click attention block is adopted for both inference and training; Discriminative affinity loss is adopted for training}
    \label{fig:framework}
\end{figure*}

For limited click influence range and click coupling issues, we present an interactive segmentation algorithm guided by local region similarity. 
The overall framework of the proposed algorithm is shown in Figure \ref{fig:framework}.

To fully expand the influence range of clicks, we introduce a click attention algorithm in Section \ref{chapter:clickattention}. 
Further, to decouple the attention coupling between positive and negative clicks, we propose a discriminative attention affinity loss in Section \ref{chapter:changeattention}.

\subsection{Preliminaries}
In a click-based interactive segmentation task, model input $x$ includes input image $x_{in}\in\mathbb{R}^{B\times3\times W\times H}$, click map $x_c\in\mathbb{R}^{B\times2\times W\times H}$, and previous segmentation results $x_{m}=\text{sigmoid}(x_P)\in\mathbb{R}^{B\times 1\times W\times H}$, where number 2 in $x_c$ represents two-channel positive and negative click maps, $B$ denotes batch size,
$x_P$ denotes the predicted target object location distribution, and $x_m$ denotes the final segmentation mask.

To balance the trade-off between efficiency and performance, Segformer \cite{NEURIPS2021_64f1f27b} is adopted as our base segmentation algorithm.
Segformer has four hierarchical Transformer encoders for feature extraction, which downsample the input $x$ at four different scales as $[1/4,1/8,1/16,1/32]$.
Each downsampling stage is denoted by $i$ with feature dimension $C_i$.
Finally, features of different scales are upsampled to a common scale and decoded by an MLP to produce the prediction result $x_P\in\mathbb{R}^{\frac{W}{4}\times\frac{H}{4}\times N_{cls}}$, where $N_{cls}=1$.

Each downsampling stage contains multiple multi-head self-attention layers, and the SRA \cite{9711179} is adopted to increase self-attention efficiency.
The attention of stage $i$, $l$-th layer and $h_i$-th head is calculated using formula (\ref{eq:selfattention1}).
\begin{equation}\label{eq:selfattention1}
    \begin{aligned}
        {A}_{i,l,h_i} = \text{softmax}\left(\frac{Q_{i,l,h_i}\cdot K_{i,l,h_i}^T}{\sqrt{{C_i}/{h_i}}}\right)
    \end{aligned}
\end{equation}
where $\cdot$ denotes the matrix multiplication, 
$i$ and $l$ denote the indexes of layer and head respectively, ${A}_{i,l,h_i}\in\mathbb{R}^{B\times L\times L}$, $Q_{i,l,h_i}\in\mathbb{R}^{B\times L\times ({C_i}/{h_i})}$ represents the query matrix, and $K_{i,l,h_i}\in\mathbb{R}^{B\times L\times ({C_i}/{h_i})}$ represents the key matrix. 
For patch $j$, its corresponding attention map is ${A}_{i,l,h_i}^j\in\mathbb{R}^{B\times 1\times L}$.
Each click is associated with a corresponding patch,
and the attention map of a click patch indicates the focus area of the network under the current click.

\subsection{Click attention based on the local regional similarity}\label{chapter:clickattention}

In this section, we propose a click attention algorithm to enhance the influence range of user clicks.

Segformer divides the input image into $L$ patches and extracts the corresponding feature $f_{i}\in\mathbb{R}^{B\times L\times C_i}$.
$f_i^j\in\mathbb{R}^{B\times 1\times C_i}$ represents the feature for patch $j$, which  often differs from other patches.
The target object is characterized by multiple patches.
Yet typically, positive clicks only fall on a limited number of patches of the target object. 
In some extreme cases, the ideal segmentation accuracy may even require clicks on all target patches. 

Inspired by this, we use $f_i^k\in\mathbb{R}^{B\times 1\times C_i}$ to denote the $k$-th positive click patch feature in stage $i$. 
The overall input feature $f_i$ and $N_k$ positive click patch features $\{f_i^k|k=1,\ldots N_k\}$ are adopted 
to calculate similarities as shown in formula (\ref{eq:similarities}).
\begin{equation}\label{eq:similarities}
 s_i^k=\text{corr}(f_i,(f_i^k)^T)   
\end{equation}
where $s_i^k\in\mathbb{R}^{B\times L\times 1}$, $\text{corr}$ denotes cross-correlation.
The final similarity score of $N_k$ positive clicks in stage $i$ can be calculated by the formula (\ref{eq:simeq}).
\begin{equation}\label{eq:simeq}
s_i=\frac{1}{N_k}\sum_{k=1}^{N_k} s_i^{k}
\end{equation}
where $s_i\in\mathbb{R}^{B\times L\times 1}$ ranges from 0 to 1. 
Next, we name $s_i$ as the \textbf{click attention} in stage $i$ for better explanation.

\begin{figure}[!t]
    \centering
    \includegraphics[scale=0.53]{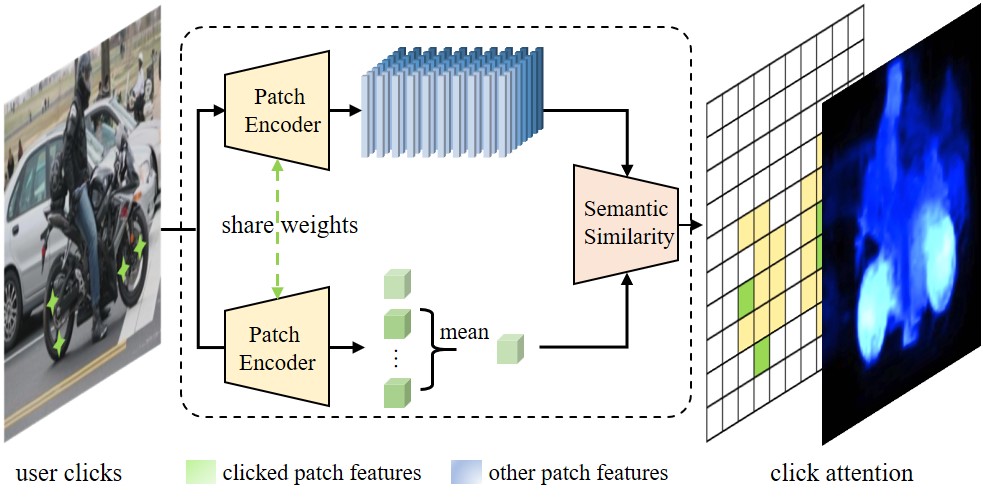}
    \caption{Click attention calculation based on patches similarity}
    \label{fig:clickattention}
\end{figure}

However, the features of different patches used for segmentation are ill-suited for similarity calculation.
Therefore, we propose a nonlinear mapping model $\phi(\cdot):\mathbb{R}^{B\times\ L\times C}\rightarrow\mathbb{R}^{B\times L\times C^\prime}$ to map patch feature $f_i$ into another space $\phi(f_i)\in\mathbb{R}^{B\times L\times C^\prime}$ for similarity calculation.
This decouples the similarity task from the segmentation task and enhances the robustness of the algorithm.
Its structure is shown in Figure \ref{fig:clickattention}.

Building on this nonlinear mapping $\phi$ (two MLP layers), we propose to use segmentation mask labels $y_i$ as supervision information for similarity learning.
In addition, the mean square loss function is adopted for training, as shown in formula (\ref{eq:simloss}).
\begin{equation}\label{eq:simloss}
    \mathcal{L}_{click}=\frac{1}{N_i}\sum_{i=1}^{N_i}\left(\text{corr}(\phi(f_i),\phi(f_i^{click}))-y_i\right)^2
\end{equation}
where $N_i$ denotes the number of downsampling stages.

To encourage the segmentation network to pay more attention to similar regions in input $x$, the click attention $s_i$ is used to weight the attention $A_{i,l,h_i}$, which can directly guide the network to focus on regions with higher similarity score, as shown in formula (\ref{eq:attnf}).
\begin{equation}\label{eq:attnf}
 \hat{A}_{i,l,h_i}=s_i\odot \bar{A}_{i,l,h_i}
\end{equation}
where $\odot$ denotes element-wise multiplication, $\bar{A}_{i,l,h_i}$ denotes the attention before softmax operation.

By adopting click attention in the attention map, regions in $x$  with a higher similarity will be enhanced, while regions with lower similarity will be suppressed, thereby guiding the network to strengthen attention to the target object.

\subsection{Discriminative attention affinity loss}\label{chapter:changeattention}
As illustrated in Figure \ref{fig:teaser}, the coupling between positive and negative clicks can activate irrelevant regions in the attention map.
The attention map $A_{i,l,h_i}$ records the attention scores among all patches, reflecting the influence of different patches on the overall attention area of the network.
We hypothesize that the coupling issue arises from overlooking the category of clicked patches in current methods.
As a result, the model struggles to learn clear discriminative relationships between positive and negative patches, leading to unnecessary context dependencies.

Previously, click attention considers solely positively-clicked patches due to the varying semantics of negative clicking patches, which may cause the network to mistakenly focus on similar objects in the background.

\begin{figure}[!h]
    \centering
    \includegraphics[scale=0.52]{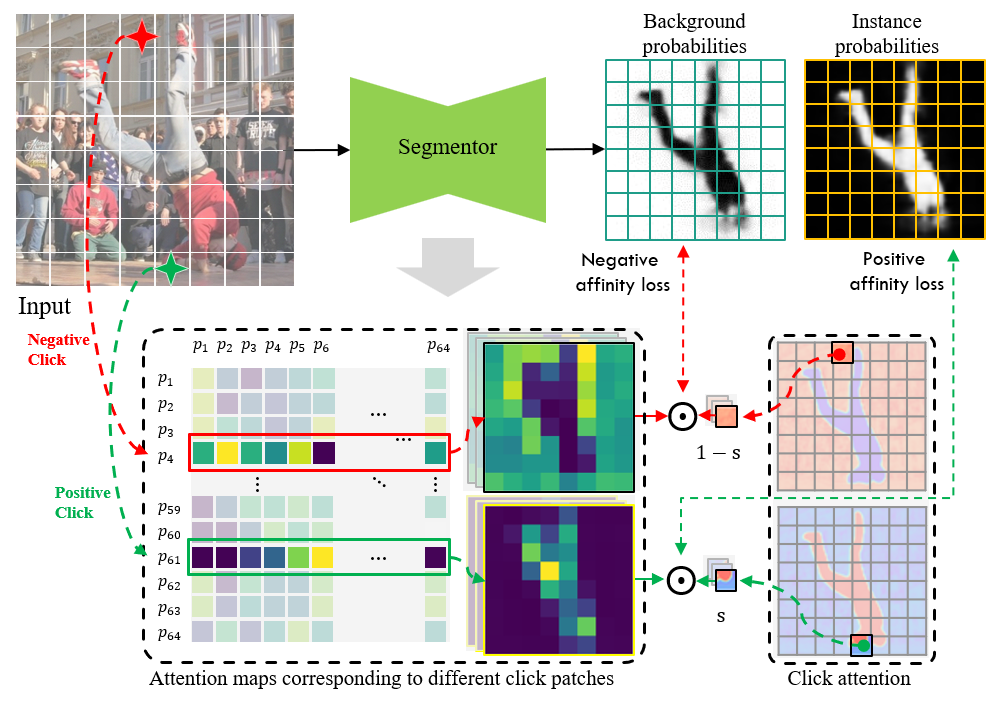}
    \caption{Illustration of discriminative affinity loss calculation based on click attention}
    \label{fig:affinityloss}
\end{figure}

To address these issues, we propose a discriminative attention affinity loss based on click attention.
By leveraging positive and negative clicks, the method can mitigate erroneous attention activation and the coupling issue mentioned earlier.
Its structure is illustrated in Figure \ref{fig:affinityloss}.

Therein, the input image is divided into $L$ patches,
$x_P$ is the target object probability,
and $x_{\hat{P}}=1-x_P$ represents the background probability. 
The $j$-th row denotes the attention map $A_{i,l,h_i}^j$ between the $j$-th patch and other patches, where $j$ is the index of the clicked patch. 

Specifically, we average the attention maps $A_{i,l,h_i}$ of multi-heads and multi-layers to calculate the attention map $A_i$ of stage $i$, as shown in formula (\ref{eq:attnmap}).
\begin{equation}\label{eq:attnmap}
    A_{i}=\frac{1}{N_{l}\times N_{h_i}}\sum_{l=1}^{N_l}\sum_{h_i=1}^{N_{h_i}} A_{i,l,h_i}
\end{equation}
where $A_{i}\in{\mathbb{R}^{B\times L\times L}}$, $N_l$ denotes the number of layers, $N_{h_i}$ denotes the number of heads. 

The attention for target object patches is $A_i\odot s_i\in R^{B\times L\times L}$, while the attention for all background patches is $A_i\odot (1-s_i)\in R^{B\times L\times L}$, where $1-s_i$ denotes the click attention between negatively-clicked patches and background patches, and $\odot$ denotes the element-wise multiplication across the second dimension of $A^{i}$.

To improve the distinction between positive and negative clicks, we calculate the affinity loss between $A_i\odot s_i$ (and $A_i\odot (1-s_i)$) and the predicted target object distributions $x_P$ (and background $x_{\hat{P}}$).
By encouraging the target object patches' attention to be more relevant to target object distribution, while encouraging the background patches' attention to be more relevant to background distribution, the coupling issue can be well alleviated.

The discriminative affinity loss is illustrated in the formula (\ref{eq:disaffloss}).
\begin{equation}\label{eq:disaffloss}
\begin{split}
    \mathcal{L}_{aff}=&\frac{1}{N_s}\sum_i^{N_s}\left\|Y_i^{pos}-x_{P}^i\odot s_i\right\|_1\\
    &+\frac{1}{N_s}\sum_i^{N_s}\left\|Y_i^{neg}-x_{\hat{P}}^i\odot (1-s_i)\right\|_1
\end{split}
\end{equation}
where $\|\cdot\|1$ denotes the $\ell_1$ loss, $N_s$ denotes the number of stages, 
and $x^{i}_P,x_{\hat{P}}^{i} \in R^{B\times 1\times L}$ are interpolated from $x_P, x_{\hat{P}}$ respectively, in stage $i$.
$x_P^i \odot s_i\in R^{B\times 1 \times L}, x_{\hat{P}}^i \odot (1-s_i)\in R^{B\times 1 \times L}$ represent predicted target object patches and background patches respectively. 
$Y^{pos}_i$ and $Y^{neg}_i$ represent the relevance of attention and prediction results, as illustrated in the formula (\ref{eq:affY}).
\begin{equation}\label{eq:affY}
\begin{split}
    Y^{pos}_i&=(A_i\odot s) \otimes x_P^\prime\\
    Y^{neg}_i&=\left(A_i\odot (1-s)\right)\otimes x_{\hat{P}}^\prime
\end{split}
\end{equation}
where $\otimes$ denotes the inner product. 
$Y^{pos}_i$ and $Y^{neg}_i$ represent the predicted probability of the target object and background distributions, respectively.

\section{Experiments}
In this chapter, we first introduce the basic configuration of our algorithm and the training/validation protocols for click-based interactive segmentation.
For ease of description, we refer to the proposed algorithm as ClickAttention in the following sections.
Next, we evaluate ClickAttention's performance by comparing it with current SOTA algorithms on widely used benchmarks, including Grabcut \cite{10.1145/1015706.1015720}, Berkeley \cite{MCGUINNESS2010434}, DAVIS \cite{perazzi2016benchmark}, SBD \cite{hariharan2011semantic},
COCO-MVal \cite{10.1007/978-3-319-10602-1_48},
PascalVOC \cite{everingham2009pascal},
DAVIS-585 \cite{9878519}.
Subsequently, an ablation study is conducted to verify the effectiveness of ClickAttention.

\subsection{Experimental Configuration}
\textbf{Model selection.} To meet the demands of high efficiency and accuracy, B0-S2, and B3-S2 in Segformer \cite{NEURIPS2021_64f1f27b} are utilized as base segmentation models. 
As illustrated in Table \ref{tab:speed_tab}, the parameter scale of Segformer is relatively small.

\textbf{Training protocol.}
The random cropping strategy is implemented to produce $512\times 512$-sized training samples, and the training procedure is conducted in an end-to-end manner.

For click simulation, the iterative learning \cite{9897365} and the training sample generation methods \cite{7780416} are adopted.
The maximum number of click points is 24, with a decaying probability of 0.8\cite{9878519}.

All algorithms in comparisons were trained on COCO \cite{10.1007/978-3-319-10602-1_48} and LVIS \cite{8954457} datasets with random flipping and resizing used for data augmentation.
The AdamW optimizer was used with $\beta_1=0.9,\beta_2=0.999$. 
Each epoch consisted of 30,000 training samples, and a total of 230 epochs were completed. 
The initial learning rate was $5\times 10^{-3}$, and it was reduced by 10 times at epochs 190 and 220. 
Our model was trained on 6 RTX 3090 GPUs and took about 24 hours.

\textbf{Evaluation protocol}
To ensure a fair comparison, we follow the performance evaluation strategies in \cite{9709986,8953578,9157109,9156403,9897365,7780416}.
In this case, each click point was sampled from the predicted region with maximum error to ensure it reached the expected ground-truth IOU (Intersection Over Union) after multiple clicks, or the click times required to reach the upper bound.

The NoC (Number of Clicks) is adopted for performance evaluation, which represents the average number of clicks required to achieve the target IOU.
The upper bound of click times is 20, and exceeding this limit means the current task has failed. 
Additionally, NoF IOU (Number of Failures) \cite{9878519} is used to evaluate the average number of failures.

\subsection{Comparison with State-of-the-Art}\label{chapter:comparesota}

\textbf{Computation analysis.}
Efficiency is an important performance indicator for interactive segmentation. To this end, we compared the differences in running speed and parameter size between ClickAttention and current SOTA algorithms, as shown in Table \ref{tab:speed_tab}.

\begin{table}[!h]
\setlength{\tabcolsep}{3pt}
\centering
\caption{Efficiency comparison with SOTA. '400', '448', and '1024' etc. denote the default input size required by different models. The speed is measured on a CPU laptop with 2.4 GHz, 4×Intel Core i5}
\label{tab:speed_tab}
\scalebox{0.9}{
\begin{tabular}{cccc}
\hline
Model Type & Params(M) & FLOPs(G) & Speed(ms) \\ \hline
RITM-hrnet32-400\cite{9897365} & 30.95 & 41.56 & 1387 \\
FirstClick-resnet101-512\cite{9157109} & 50.37 & 216 & 6267 \\
SimpleClick-ViT-B-448\cite{liu2022simpleclick} & 84.89 & 96.46 &  8640\\
SimpleClick-ViT-L-448\cite{liu2022simpleclick} & 322.18 & 266.44 & 10025 \\
HQ-SAM-ViT-B-1024\cite{sam_hq} & 637.23 & 2830.34 & 11238 \\
SAM-ViT-H-1024\cite{kirillov2023segment} & 635.63 & 2802.69 & 10351 \\
FocalClick-B3-S2-256\cite{9878519} & 45.66 & 12.37 & {706} \\ \hline
Ours-B0-S2 & \textbf{4.36}$_{\textcolor{red}{\downarrow{90\%}}}$ & \textbf{2.11}$_{\textcolor{red}{\downarrow{83\%}}}$ & \textbf{220}$_{\textcolor{red}{\downarrow{68\%}}}$ \\ 
Ours-B3-S2 & 46.44 & 12.43 & 729 \\ \hline
\end{tabular}
}
\end{table}

The model parameters of the current SOTA algorithm SimpleClick are 84.89e16 and 322.18e16, which are 1.82 times and 6.94 times larger than our largest model, respectively. Due to its larger input size, it takes more than 10 times longer for the algorithm in question to run compared to our algorithm. Additionally, both RITM and FirstClick-resnet101 have a larger number of parameters than the algorithm proposed in this paper, with FirstClick-resnet101 taking 8.6 times longer to run than our algorithm. Except for FocalClick, all models have a runtime exceeding 1 second. Due to the larger input size of SimpleClick, its inference time is more than ten times that of our algorithm. In addition, the parameter sizes of RITM and FirstClick-resnet101 are also larger than the proposed algorithm, with the inference time of FirstClick-resnet101 being 8.6 times as long as ours. Except for FocalClick, all compared models have more than 1-second running speed. Interactive segmentation algorithms are widely used in crowdsourcing annotation tasks, where device performance is typically low. Therefore, a large number of parameters can create a significant burden on devices with only a CPU, leading to reduced work efficiency.

Furthermore, our ClickAttention model has significantly fewer parameters compared to the Segment Anything Model (SAM) and HQ-SAM, with only 1/13.69 and 1/13.72 of their parameter sizes, respectively. 
It's worth noting that their larger input size (1024x1024) resulted in FLOPs that are 225 and 227 times larger than ours, respectively.
As a result, these methods may be difficult to apply in worldwide crowdsourcing image annotation scenarios (most of the devices are low-performance).
However, SAM is an excellent approach that only encodes input images once and then processes the clicks in its transformer decoder head. 
This helps SAM (and HQ-SAM) maintain a fast inference speed even with large FLOPs.
This is highly inspiring for our future research.

In comparison, the Segformer-based FocalClick and ClickAttention demonstrate faster inference time. Compared to FocalClick, ClickAttention achieved better performance at the cost of a limited loss in speed.

\begin{table*}[!t]
\setlength{\tabcolsep}{3pt}
\centering
\caption{Evaluation results on GrabCut, Berkeley, SBD, DAVIS, COCO Mval and Pascal VOC datasets. 'NoC 85/90' denotes the average Number of Clicks required to get IoU of 85/90\%. All methods are trained on COCO \cite{10.1007/978-3-319-10602-1_48} and LVIS \cite{8954457} datasets (except SAM and HQSAM, more details can be found in chapter \ref{chapter:comparesota}). \textcolor{red}{Red arrow}: ClickAttention $>$ 2n place. \textcolor{blue}{Blue arrow}: ClickAttention $>$ 3rd place. \textbf{Bold font}: best performance}
\label{tab:sotacompar}
\scalebox{0.93}{
\begin{tabular}{cccccccccc}
\hline
 \multirow{2}{*}{Method} & GrabCut\cite{10.1145/1015706.1015720} & Berkeley\cite{MCGUINNESS2010434} & \multicolumn{2}{c}{SBD\cite{hariharan2011semantic}} & DAVIS\cite{perazzi2016benchmark} & \multicolumn{2}{c}{COCO\_MVal\cite{10.1007/978-3-319-10602-1_48}} & \multicolumn{2}{c}{PascalVOC\cite{everingham2009pascal}} \\
 & NoC 90 & NoC 90 & NoC 85 & NoC 90 & NoC 90 & NoC 85 & NoC 90 & NoC 85 & NoC 90 \\ \hline
f-BRS-B-hrnet32\cite{9156403} & 1.69 & 2.44 & 4.37 & 7.26 & 6.50 & - & - & -& - \\
RITM-hrnet18s\cite{9897365} & 1.68 & 2.60 & 4.25 & 6.84 & 5.98 & - & 3.58 & 2.57 & - \\
RITM-hrnet32\cite{9897365} & 1.56 & 2.10 & 3.59 & 5.71 & 5.34 & 2.18 & 3.03 & 2.21 & 2.59 \\
EdgeFlow-hrnet18\cite{9607681} & 1.72 & 2.40 & - & - & 5.77 & - & - & - & - \\
SimpleClick-ViT-B\cite{liu2022simpleclick} & 1.50 & 1.86 & 3.38 & 5.50 & 5.18 & 2.18 & 2.92 & 2.06 & 2.38 \\
SimpleClick-ViT-L\cite{liu2022simpleclick} & {1.46} & 1.83 & \textbf{2.92} & \textbf{4.85} & 4.77 & \textbf{1.96} & \textbf{2.63} & \textbf{1.71} & \textbf{1.93} \\ 
SAM-ViT-L\cite{sam_hq} & 1.62 & 2.25 & 5.98 & 9.63 & 6.21 & 3.46 & 5.60 & 2.20 & 2.68\\ 
HQ-SAM-ViT-L\cite{liu2022simpleclick} & 1.84 & 2.00 & 6.23 & 9.66 & 5.58 & 3.81 & 5.94 & 2.50 & 2.93\\ 
FocalClick-B0-S2\cite{9878519} & 1.90 & 2.92 & 5.14 & 7.80 & 6.47 & 3.23 & 4.37 & 3.55 & 4.24 \\
FocalClick-B3-S2\cite{9878519} & 1.68 & \textbf{1.71} & 3.73 & 5.92 & 5.59 & 2.45 & 3.33 & 2.53 & 2.97 \\\hline
Ours-B0-S2 & 1.66 & 2.65 & 4.45 & 6.90 & 5.44 & 2.81 & 3.77 & 2.87 & 3.36 \\
Ours-B3-S2 & 1.48 & 1.89 & 3.18$_{\textcolor{blue}{\uparrow{5.9\%}}}$ & 5.17$_{\textcolor{blue}{\uparrow{6.2\%}}}$ & \textbf{4.67}$_{\textcolor{red}{\uparrow{2.0\%}}}$ & 2.05$_{\textcolor{blue}{\uparrow{5.9\%}}}$ & 2.83$_{\textcolor{blue}{\uparrow{3.0\%}}}$ & 1.96$_{\textcolor{blue}{\uparrow{4.8\%}}}$ & 2.26$_{\textcolor{blue}{\uparrow{5.0\%}}}$ \\ \hline
\end{tabular}
}
\end{table*}
\textbf{Performance on existing benchmarks}
Performance on existing benchmarks Classical methods such as Graph Cut  \cite{boykov2001interactive} and Geodesic star convexity \cite{gulshan2010geodesic} were not compared in the study due to their performance disadvantage. The performance comparison results of ClickAttention and current SOTA on mainstream benchmarks are shown in Table \ref{tab:sotacompar}.

As illustrated in Table 1, Ours-B0-S2 achieved better performance on multiple datasets compared to FocalClick-B3-S2 \cite{9878519}, which also uses Segformer. In detail, Ours-B0-S2 performed 1.2\% and 2.7\% better than FocalClick-B3-S2 on GrabCut and DAVIS, respectively. This indicates that the proposed algorithm still has competitive performance with fewer parameters. We believe the reason for this is that the proposed algorithm fully utilizes all user clicks, whereas FocalClick's target crop operation discards points outside the target region, causing it to only refine the local regions of large targets, but ignore the overall accuracy. 

In addition, compared to FocalClick-B3-S2, Ours-B3-S2 showed better performance in almost all benchmarks. As shown in Table 1, the algorithms compared performed more poorly on the SBD and DAVIS datasets. This is likely because SBD contains over 6000 test samples, and DAVIS includes a significant number of challenging cases, resulting in relatively high values of NoC90. Ours-B3-S2 achieved the best performance on DAVIS, with a 1\% improvement over the second-place algorithm SimpleClick\cite{liu2022simpleclick}. On SBD, our algorithm's NoC90 improved by 4.9\% compared to the third-place algorithm. Combined with Table 2, the results show that our algorithm achieves competitive performance with fewer parameters. On GrabCut, Ours-B3-S2 and SimpleClick achieved the best NoC90 performance. Except for ranking fourth on the Berkeley benchmark, Ours-B3-S2's performance on other datasets is only second to SimpleClick. 
After analyzing the characteristics of the Berkeley dataset, we found that it has a low resolution (480x320) and there are many cases of intricate edges, including antlers and feathers. The performance of the proposed algorithm still needs to be improved under these conditions.

Compared to both SAM and HQ-SAM, the proposed algorithm also achieved better performance with fewer parameters. It is worth noting that the SAM is trained on a large-scale dataset containing 10e9 well-annotated images \cite{kirillov2023segment}. 
In comparison, the proposed algorithm only requires the COCO-LIVIS dataset (containing 10e4 well-annotated images) for training. This indicates that our algorithm has higher efficiency.

\subsection{Performance for Mask Correction}
The Mask Correction task involves refining the provided initial mask with an IOU value between 0.75 and 0.85.
We evaluated the effectiveness of ClickAttention in Mask Correction using the DAVIS-585 dataset \cite{9878519}. The experimental results are shown in Table \ref{tab:maskcorrection}.

\begin{figure*}[!t]
    \centering
    \includegraphics[width=6.3in]{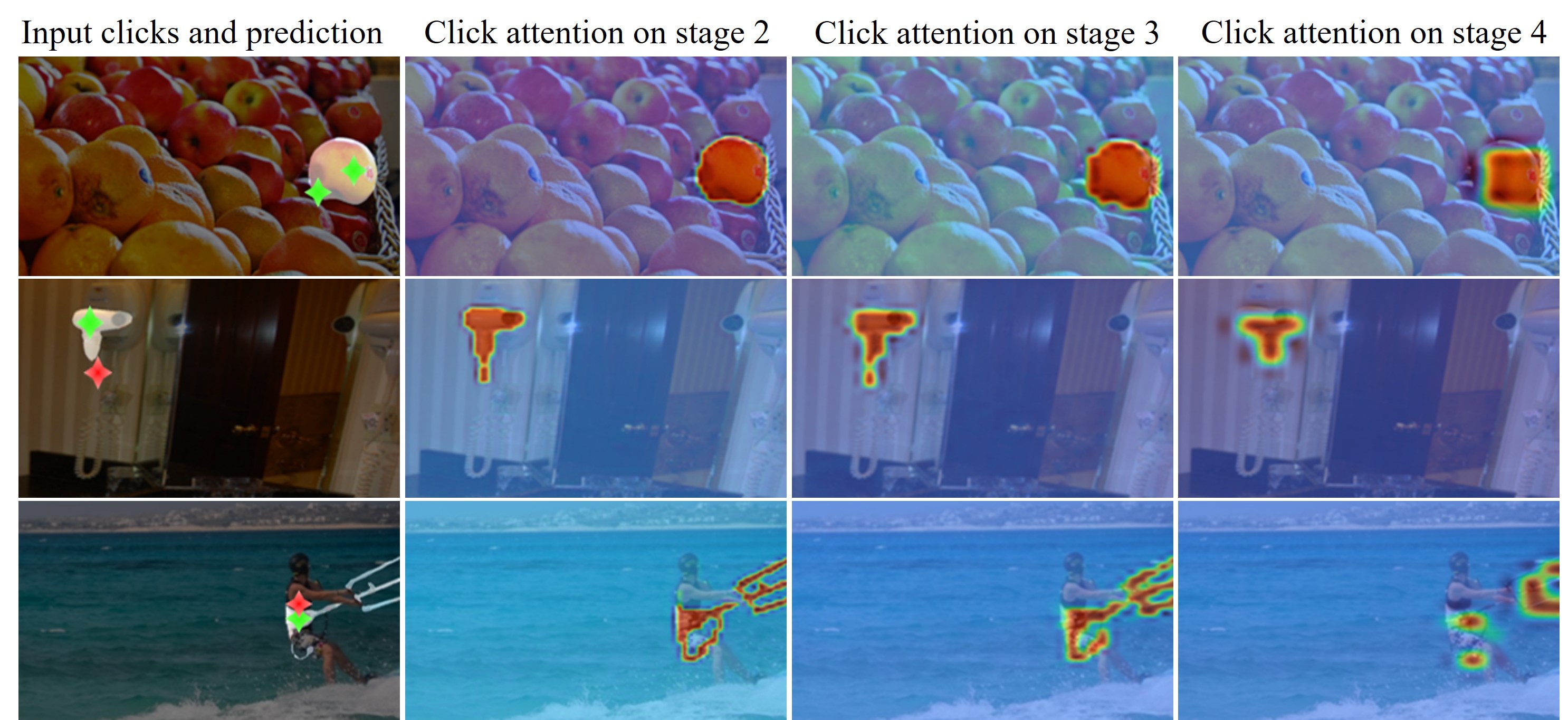}
    \caption{Qualitative results for the effectiveness of the click attention. The first column shows the input clicks and prediction results. Thee remaining columns display the click attention in different downsampling stages, which demonstrates its ability to capture the focus range of positive clicks}
    \label{fig:attnW}
\end{figure*}

\begin{table}[!h]
\setlength{\tabcolsep}{3pt}
\centering
\caption{Quantitative results on DAVIS-585 benchmark. The metrics ‘NoC’ and ‘NoF’ mean the average Number of Clicks required and the Number of Failure examples for the target IOU. All models are trained on COCO\cite{10.1007/978-3-319-10602-1_48}+LVIS\cite{8954457}}
\label{tab:maskcorrection}
\scalebox{0.7}{
\begin{tabular}{cccc|ccc}
\hline
\multirow{2}{*}{Method} & \multicolumn{3}{c|}{DAVIS585-SP} & \multicolumn{3}{c}{DAVIS585-ZERO} \\ 
 & NoC85 & NoC90 & NoF85 & NoC85 & NoC90 & NoF85 \\ \hline
RITM-hrnet18s\cite{9897365} & 3.71 & 5.96 & 49 & 5.34 & 7.57 & 52 \\
RITM-hrnet32\cite{9897365} & 3.68 & 5.57 & 46 & 4.74 & 6.74 & 45 \\
SimpleClick-ViT-B\cite{liu2022simpleclick} & 2.24 & 3.10 & 25 & 4.06 & 5.83 & 42 \\
SimpleClick-ViT-L\cite{liu2022simpleclick} & 1.81 & 2.57 & 25 & \textbf{3.39} & \textbf{4.88} & 36 \\
FocalClick-hrnet32-S2\cite{9878519} & 2.32 & 3.09 & 28 & 4.77 & 6.84 & 48 \\
FocalClick-B3-S2\cite{9878519} & 2.00 & 2.76 & 22 & 4.06 & 5.89 & 43 \\ \hline
Ours-B3-S2 & \textbf{1.74}$_{\textcolor{red}{\uparrow{3.8\%}}}$ & \textbf{2.42}$_{\textcolor{red}{\uparrow{5.8\%}}}$ & \textbf{15}$_{\textcolor{red}{\uparrow{40\%}}}$ & 3.78$_{\textcolor{blue}{\uparrow{6.9\%}}}$ & 5.55$_{\textcolor{blue}{\uparrow{4.8\%}}}$ & 43 \\ \hline
\end{tabular}
}
\end{table}

It can be found that, compared to RITM, SimpleClick, and FocalClick, ClickAttention has the best performance on the DAVIS585-SP task. Compared with the second-ranked SimpleClick, we have improved by 5.8\% on NoC90, which proves the advanced nature of the algorithm in this paper on the mask correction task. Compared to FocalClick, ClickAttention improves NoC 85 and NoC 90 in mask correction by 13\% and 12.3\% respectively. Compared with the second-ranked SimpleClick, we have improved by 5.8\% NoC90 performance, which demonstrates the effectiveness of the proposed algorithm in mask correction tasks.

We believe the reason is that the input of the mask can provide certain prior information for the network. However, the existing algorithm only treats the mask as an additional modal feature and does not make full use of it. In our work, we mainly calculate the similarity between the user click area and other area features. Among these features, the prior mask provides consistent feature expression for the target area tokens. So that it helps the algorithm to better find all similar tokens. Compared to other methods, the algorithm proposed in this paper can make better use of prior mask information.

On the DAVIS585-ZERO task, ClickAttention ranks second, only behind SimpleClick-ViT-L, whose parameter size is 6.94 times larger than ours. Compared to FocalClick, the proposed algorithm improves NoC 85 and NoC 90 by 6.9\% and 5.7\% respectively. This trend is consistent with the results shown in Table \ref{tab:sotacompar}.

\subsection{Ablation Study}
We conducted ablation experiments based on the B0-S2 model on DAVIS \cite{perazzi2016benchmark} and DAVIS585-SP \cite{9878519} datasets.

\textbf{Holistic analysis.} Each component is evaluated to verify the effectiveness of the proposed algorithm, as presented in Table \ref{tab:ablation}. 
We used Segformer \cite{9878519} as the baseline, and studied the impact of performance by incorporating click attention and discriminative affinity loss, respectively.

\begin{table}[!h]
\setlength{\tabcolsep}{3pt}
\centering
\caption{Ablation studies on both interactive segmentation from scratch and interactive mask correction. 
'CA' and 'DAA' denote click attention and discriminative attention affinity, respectively }
\label{tab:ablation}
\scalebox{0.82}{
\begin{tabular}{cccc|ccc}
\hline
\multirow{2}{*}{Method} & \multicolumn{3}{c|}{DAVIS} & \multicolumn{3}{c}{DAVIS585-SP} \\
 & NoC85 & NoC90 & NoF90 & NoC85 & NoC90 & NoF90 \\ \hline
baseine-B0-S2 & 4.83 & 6.47 & 55 & 2.63 & 3.69 & 54 \\
+CA & 4.22$_{\textcolor{red}{\uparrow{12\%}}}$ & 5.65$_{\textcolor{red}{\uparrow{12\%}}}$ & 42$_{\textcolor{red}{\uparrow{23\%}}}$ & 2.23$_{\textcolor{red}{\uparrow{15\%}}}$ & 3.44$_{\textcolor{red}{\uparrow{6.7\%}}}$ & 34$_{\textcolor{red}{\uparrow{37\%}}}$ \\
+CA+DAA & 4.08$_{\textcolor{red}{\uparrow{15\%}}}$ & 5.44$_{\textcolor{red}{\uparrow{15\%}}}$ & 35$_{\textcolor{red}{\uparrow{36\%}}}$ & 2.18$_{\textcolor{red}{\uparrow{17\%}}}$ & 3.19$_{\textcolor{red}{\uparrow{13\%}}}$ & 25$_{\textcolor{red}{\uparrow{54\%}}}$ \\ \hline
\end{tabular}
}
\end{table}

As illustrated in Table \ref{tab:ablation},
CA brings significant improvements in mask correction and DAVIS datasets.
Moreover, by incorporating DAA, the performance is further enhanced in two evaluation metrics, thereby confirming the effectiveness of the proposed approach.

\textbf{Click attention analysis.}
To verify that ClickAttention learns the correct click attention based on local region similarity, we visualized the click attention of various downsampling stages in some typical scenarios, as illustrated in Figure \ref{fig:attnW}. 
To ensure the reliability of the verification, all visualization operations are performed in the evaluation process.

It is evident that with an increase in the number of stages, the resolution of click attention decreases. Nonetheless, each stage's click attention accurately captures the user's intended focus which validates the effectiveness of click attention.

\begin{figure*}[]
    \centering
    \includegraphics[width=7in]{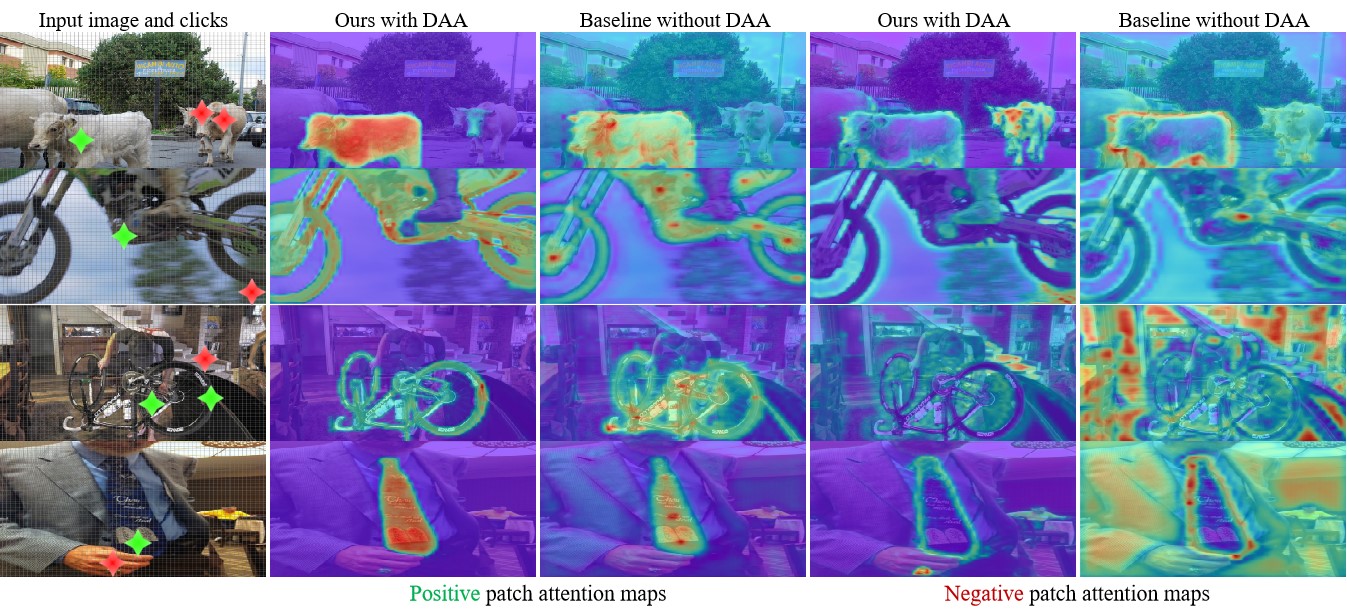}
    \caption{Qualitative results for the effectiveness of the Discriminative click Attention Affinity. The first column shows the input image and clicks. The second and third columns display the comparison of positively-clicked region attention maps between the baseline and ClickAttention. The fourth and fifth columns present the negative attention maps between the baseline and ClickAttention}
    \label{fig:DAA}
\end{figure*}

\begin{figure*}[]
    \centering
    \includegraphics[width=6.7in]{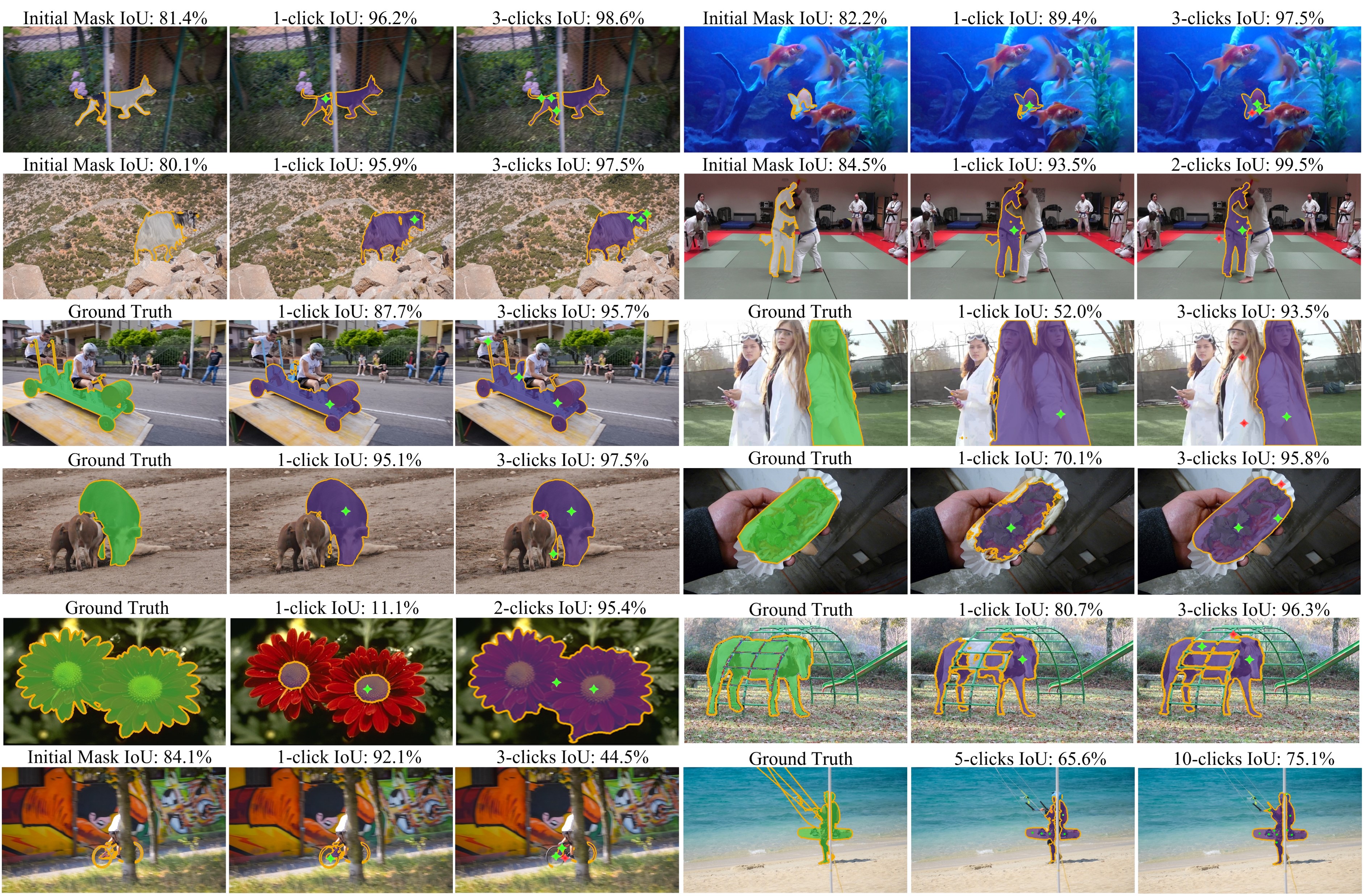}
    \caption{Qualitative results on existing benchmarks. Rows 1 and 2 present the cases where starting from initial masks. Rows 3, 4, and 5 show examples of annotation from scratch. The last row shows the worst and most challenging cases}
    \label{fig:vis_result}
\end{figure*}

\begin{figure*}[]
    \centering
    \includegraphics[width=6.7in]{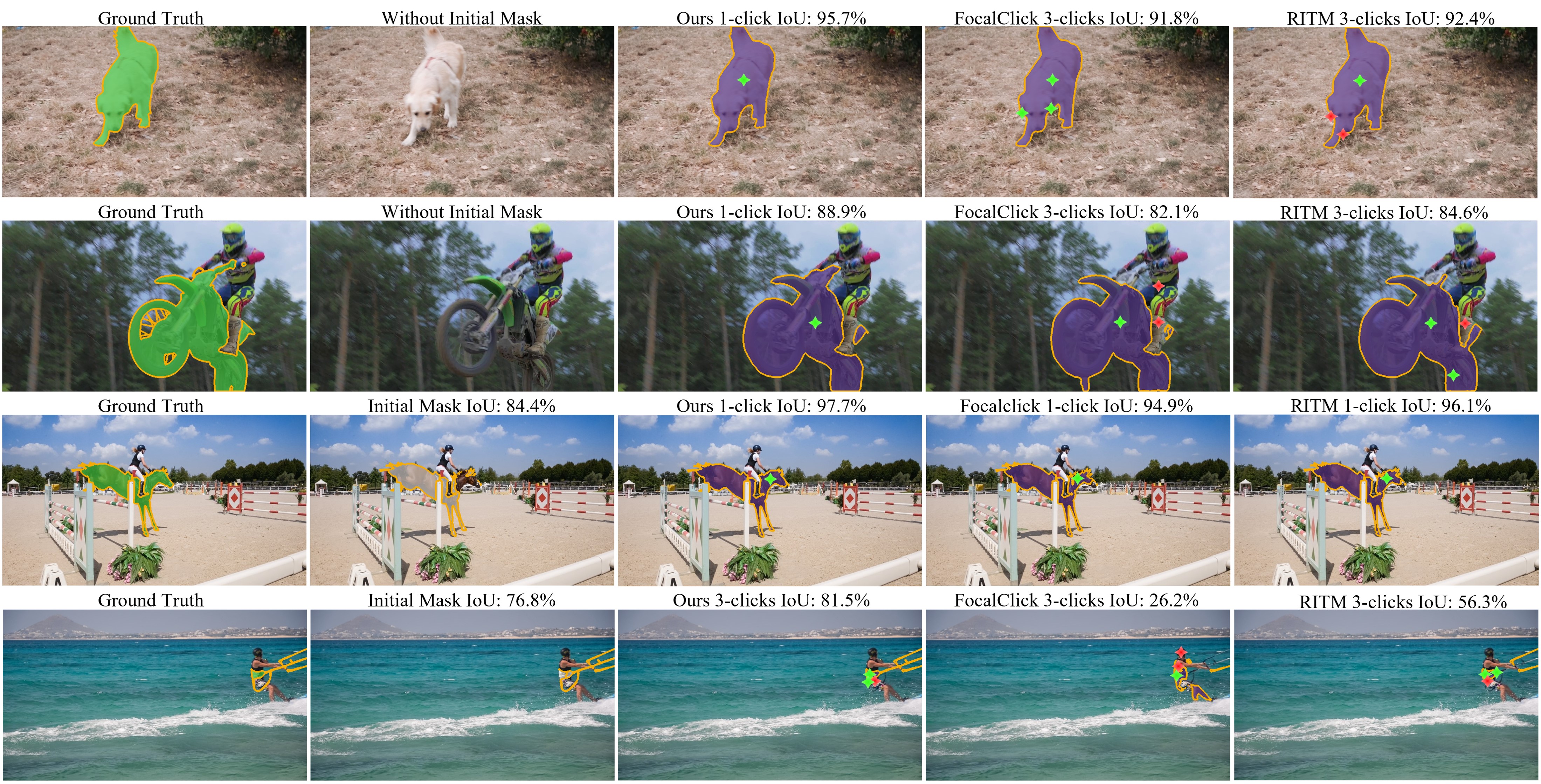}
    \caption{Comparison results on DAVIS-585 benchmarks. Rows 1 and 2 show the advantages of ClickAttention compared to the FocalClick-B3-S2 without initial masks. Rows 3 and 4 demonstrate comparisons of the mask correction task }
    \label{fig:vis_compare_baseline}
\end{figure*}

\textbf{Discriminative attention analysis}.
To verify the effectiveness of the proposed discriminative affinity loss in decoupling influences of positive and negative clicks, we conducted a visualization analysis of the positive/negative region attention maps in several typical scenarios, as shown in Figure \ref{fig:DAA}.

The third and fifth columns of Figure \ref{fig:DAA} demonstrate that the positively and negative-clicked regions' attention of the baseline model tends to confuse the target object with the background, which in turn impacts the performance of training and inference.

After incorporating the DAA, as shown in columns 2 and 4, positive clicks can focus on the target effectively, while the negatively-clicked region attention is also able to effectively decouple from the positively-clicked region attention. 
As a result, negative clicks can concentrate exclusively on the parts where the background and the target object are easily confused.

It is worth noting that the attention of negatively-clicked regions is successfully focused on the edges of the target object, providing valuable guidance for the model's fine-grained segmentation.

\subsection{Qualitative Result}

To verify the effectiveness of ClickAttention, we evaluated its precision changes under multiple clicks on several typical datasets. 
The results are shown in Figure \ref{fig:vis_result}.

The first two rows of Figure \ref{fig:vis_result} demonstrate the performance of ClickAttention in mask correction.
It can be observed that ClickAttention exhibits excellent performance and achieves high accuracy within a limited number of clicks.
The following three rows represent the performance of ClickAttention without an initial mask, showing that ClickAttention performs well under a single click, and the accuracy increases significantly after three clicks. However, the last row in Figure 8 illustrates some of the worst and most challenging cases by using ClickAttention, which demonstrates its limitations. Specifically, the segmentation of tiny and delicate targets requires multiple clicks to achieve better results. Similarly, target-obstructing situations pose challenges for ClickAttention. In some extreme cases, ClickAttention may produce poor or even failed segmentation results.

To further verify the effectiveness of ClickAttention, we compared it with SOTA algorithms including FocalClick \cite{9878519} and RITM \cite{9897365} under the DAVIS585 benchmarks.
The corresponding quantitative analysis results are shown in Figure \ref{fig:vis_compare_baseline}.

As shown in Figure \ref{fig:vis_compare_baseline}, without initial masks, rows 1 and 2 demonstrate that one-clicked ClickAttention obtains better results than 3 times clicked FocalClick \cite{9878519} and RITM \cite{9897365}. 
Rows 3 and 4 show that in the task of mask correction, ClickAttention outperforms FocalClick \cite{9878519} and RITM \cite{9897365} after once corrected click. 

\section{Conclusion}
The current paper investigates the relationship between user interaction and segmentation models in interactive segmentation.
Specifically, to address the issue of small click influence range, we propose a click attention algorithm that expands the attention range based on the similarity between positively-clicked regions and overall input. Additionally, to address the coupling issue, we further propose a discriminative affinity loss that effectively utilizes the discriminability between positive and negative clicks. Our extensive experimental results demonstrate that our proposed method significantly improves algorithm performance while achieving better performance with smaller parameter scales. 
Compared to current state-of-the-art algorithms, our proposed algorithm shows competitive performance.

\bibliographystyle{ieeetr}
\bibliography{newbib.bib}

\begin{thebibliography}{10}

\bibitem{dosovitskiy2020image}
A.~Dosovitskiy, L.~Beyer, A.~Kolesnikov, D.~Weissenborn, X.~Zhai,
  T.~Unterthiner, M.~Dehghani, M.~Minderer, G.~Heigold, S.~Gelly, {\em et~al.},
  ``An image is worth 16x16 words: Transformers for image recognition at
  scale,'' {\em arXiv preprint arXiv:2010.11929}, 2020.

\bibitem{9878519}
X.~Chen, Z.~Zhao, Y.~Zhang, M.~Duan, D.~Qi, and H.~Zhao, ``Focalclick: Towards
  practical interactive image segmentation,'' in {\em 2022 IEEE/CVF Conference
  on Computer Vision and Pattern Recognition (CVPR)}, pp.~1290--1299, 2022.

\bibitem{liu2022simpleclick}
Q.~Liu, Z.~Xu, G.~Bertasius, and M.~Niethammer, ``Simpleclick: Interactive
  image segmentation with simple vision transformers,'' {\em arXiv preprint
  arXiv:2210.11006}, 2022.

\bibitem{7982754}
K.~Kim and S.-W. Jung, ``Interactive image segmentation using semi-transparent
  wearable glasses,'' {\em IEEE Transactions on Multimedia}, vol.~20, no.~1,
  pp.~208--223, 2018.

\bibitem{7108034}
K.~Li and W.~Tao, ``Adaptive optimal shape prior for easy interactive object
  segmentation,'' {\em IEEE Transactions on Multimedia}, vol.~17, no.~7,
  pp.~994--1005, 2015.

\bibitem{7780416}
N.~Xu, B.~Price, S.~Cohen, J.~Yang, and T.~Huang, ``Deep interactive object
  selection,'' in {\em 2016 IEEE Conference on Computer Vision and Pattern
  Recognition (CVPR)}, pp.~373--381, 2016.

\bibitem{9187581}
T.~Wang, Z.~Ji, J.~Yang, Q.~Sun, and P.~Fu, ``Global manifold learning for
  interactive image segmentation,'' {\em IEEE Transactions on Multimedia},
  vol.~23, pp.~3239--3249, 2021.

\bibitem{7544505}
T.~Wang, Z.~Ji, Q.~Sun, Q.~Chen, and X.-Y. Jing, ``Interactive multilabel image
  segmentation via robust multilayer graph constraints,'' {\em IEEE
  Transactions on Multimedia}, vol.~18, no.~12, pp.~2358--2371, 2016.

\bibitem{9157109}
Z.~Lin, Z.~Zhang, L.-Z. Chen, M.-M. Cheng, and S.-P. Lu, ``Interactive image
  segmentation with first click attention,'' in {\em 2020 IEEE/CVF Conference
  on Computer Vision and Pattern Recognition (CVPR)}, pp.~13336--13345, 2020.

\bibitem{DBLP:journals/corr/abs-2003-07932}
M.~Forte, B.~L. Price, S.~Cohen, N.~Xu, and F.~Piti{\'{e}}, ``Getting to 99{\%}
  accuracy in interactive segmentation,'' {\em CoRR}, vol.~abs/2003.07932,
  2020.

\bibitem{9156403}
K.~Sofiiuk, I.~Petrov, O.~Barinova, and A.~Konushin, ``F-brs: Rethinking
  backpropagating refinement for interactive segmentation,'' in {\em 2020
  IEEE/CVF Conference on Computer Vision and Pattern Recognition (CVPR)},
  pp.~8620--8629, 2020.

\bibitem{9607681}
Y.~Hao, Y.~Liu, Z.~Wu, L.~Han, Y.~Chen, G.~Chen, L.~Chu, S.~Tang, Z.~Yu,
  Z.~Chen, and B.~Lai, ``Edgeflow: Achieving practical interactive segmentation
  with edge-guided flow,'' in {\em 2021 IEEE/CVF International Conference on
  Computer Vision Workshops (ICCVW)}, pp.~1551--1560, 2021.

\bibitem{9709986}
X.~Chen, Z.~Zhao, F.~Yu, Y.~Zhang, and M.~Duan, ``Conditional diffusion for
  interactive segmentation,'' in {\em 2021 IEEE/CVF International Conference on
  Computer Vision (ICCV)}, pp.~7325--7334, 2021.

\bibitem{liu2022pseudoclick}
Q.~Liu, M.~Zheng, B.~Planche, S.~Karanam, T.~Chen, M.~Niethammer, and Z.~Wu,
  ``Pseudoclick: Interactive image segmentation with click imitation,'' in {\em
  Computer Vision--ECCV 2022: 17th European Conference, Tel Aviv, Israel,
  October 23--27, 2022, Proceedings, Part VI}, pp.~728--745, Springer, 2022.

\bibitem{majumder2019content}
S.~Majumder and A.~Yao, ``Content-aware multi-level guidance for interactive
  instance segmentation,'' in {\em Proceedings of the IEEE/CVF Conference on
  Computer Vision and Pattern Recognition}, pp.~11602--11611, 2019.

\bibitem{9729600}
Z.~Ding, T.~Wang, Q.~Sun, and F.~Chen, ``Rethinking click embedding for deep
  interactive image segmentation,'' {\em IEEE Transactions on Industrial
  Informatics}, vol.~19, no.~1, pp.~261--273, 2023.

\bibitem{kirillov2023segment}
A.~Kirillov, E.~Mintun, N.~Ravi, H.~Mao, C.~Rolland, L.~Gustafson, T.~Xiao,
  S.~Whitehead, A.~C. Berg, W.-Y. Lo, P.~Dollár, and R.~Girshick, ``Segment
  anything,'' 2023.

\bibitem{tancik2020fourier}
M.~Tancik, P.~P. Srinivasan, B.~Mildenhall, S.~Fridovich-Keil, N.~Raghavan,
  U.~Singhal, R.~Ramamoorthi, J.~T. Barron, and R.~Ng, ``Fourier features let
  networks learn high frequency functions in low dimensional domains,'' 2020.

\bibitem{sam_hq}
L.~Ke, M.~Ye, M.~Danelljan, Y.~Liu, Y.-W. Tai, C.-K. Tang, and F.~Yu, ``Segment
  anything in high quality,'' {\em arXiv:2306.01567}, 2023.

\bibitem{chen2023sam}
T.~Chen, L.~Zhu, C.~Ding, R.~Cao, S.~Zhang, Y.~Wang, Z.~Li, L.~Sun, P.~Mao, and
  Y.~Zang, ``Sam fails to segment anything? -- sam-adapter: Adapting sam in
  underperformed scenes: Camouflage, shadow, and more,'' 2023.

\bibitem{zhao2023fast}
X.~Zhao, W.~Ding, Y.~An, Y.~Du, T.~Yu, M.~Li, M.~Tang, and J.~Wang, ``Fast
  segment anything,'' 2023.

\bibitem{10132428}
J.~Ma, Y.~Bai, B.~Zhong, W.~Zhang, T.~Yao, and T.~Mei, ``Visualizing and
  understanding patch interactions in vision transformer,'' {\em IEEE
  Transactions on Neural Networks and Learning Systems}, pp.~1--10, 2023.

\bibitem{su2022sasformer}
H.~Su, Y.~Ye, W.~Hua, L.~Cheng, and M.~Song, ``Sasformer: Transformers for
  sparsely annotated semantic segmentation,'' {\em arXiv preprint
  arXiv:2212.02019}, 2022.

\bibitem{10.1145/1015706.1015720}
C.~Rother, V.~Kolmogorov, and A.~Blake, ``"grabcut": Interactive foreground
  extraction using iterated graph cuts,'' {\em ACM Trans. Graph.}, vol.~23,
  p.~309–314, aug 2004.

\bibitem{MCGUINNESS2010434}
K.~McGuinness and N.~E. O’Connor, ``A comparative evaluation of interactive
  segmentation algorithms,'' {\em Pattern Recognition}, vol.~43, no.~2,
  pp.~434--444, 2010.
\newblock Interactive Imaging and Vision.

\bibitem{perazzi2016benchmark}
F.~Perazzi, J.~Pont-Tuset, B.~McWilliams, L.~Van~Gool, M.~Gross, and
  A.~Sorkine-Hornung, ``A benchmark dataset and evaluation methodology for
  video object segmentation,'' in {\em Proceedings of the IEEE conference on
  computer vision and pattern recognition}, pp.~724--732, 2016.

\bibitem{hariharan2011semantic}
B.~Hariharan, P.~Arbel{\'a}ez, L.~Bourdev, S.~Maji, and J.~Malik, ``Semantic
  contours from inverse detectors,'' in {\em 2011 international conference on
  computer vision}, pp.~991--998, IEEE, 2011.

\bibitem{10.1007/978-3-319-10602-1_48}
T.-Y. Lin, M.~Maire, S.~Belongie, J.~Hays, P.~Perona, D.~Ramanan,
  P.~Doll{\'a}r, and C.~L. Zitnick, ``Microsoft coco: Common objects in
  context,'' in {\em Computer Vision -- ECCV 2014} (D.~Fleet, T.~Pajdla,
  B.~Schiele, and T.~Tuytelaars, eds.), (Cham), pp.~740--755, Springer
  International Publishing, 2014.

\bibitem{everingham2009pascal}
M.~Everingham, L.~Van~Gool, C.~K. Williams, J.~Winn, and A.~Zisserman, ``The
  pascal visual object classes (voc) challenge,'' {\em International journal of
  computer vision}, vol.~88, pp.~303--308, 2009.

\bibitem{1704833}
L.~Grady, ``Random walks for image segmentation,'' {\em IEEE Transactions on
  Pattern Analysis and Machine Intelligence}, vol.~28, no.~11, pp.~1768--1783,
  2006.

\bibitem{gulshan2010geodesic}
V.~Gulshan, C.~Rother, A.~Criminisi, A.~Blake, and A.~Zisserman, ``Geodesic
  star convexity for interactive image segmentation,'' in {\em 2010 IEEE
  Computer Society Conference on Computer Vision and Pattern Recognition},
  pp.~3129--3136, IEEE, 2010.

\bibitem{kim2010nonparametric}
T.~H. Kim, K.~M. Lee, and S.~U. Lee, ``Nonparametric higher-order learning for
  interactive segmentation,'' in {\em 2010 IEEE computer society conference on
  computer vision and pattern recognition}, pp.~3201--3208, IEEE, 2010.

\bibitem{Zhang2022IntentionawareFP}
C.~Zhang, C.~Hu, Y.~Liu, and X.~He, ``Intention-aware feature propagation
  network for interactive segmentation,'' {\em ArXiv}, vol.~abs/2203.05145,
  2022.

\bibitem{8953578}
W.-D. Jang and C.-S. Kim, ``Interactive image segmentation via backpropagating
  refinement scheme,'' in {\em 2019 IEEE/CVF Conference on Computer Vision and
  Pattern Recognition (CVPR)}, pp.~5292--5301, 2019.

\bibitem{9897365}
K.~Sofiiuk, I.~A. Petrov, and A.~Konushin, ``Reviving iterative training with
  mask guidance for interactive segmentation,'' in {\em 2022 IEEE International
  Conference on Image Processing (ICIP)}, pp.~3141--3145, 2022.

\bibitem{DBLP:journals/corr/abs-2111-06377}
K.~He, X.~Chen, S.~Xie, Y.~Li, P.~Doll{\'{a}}r, and R.~B. Girshick, ``Masked
  autoencoders are scalable vision learners,'' {\em CoRR}, vol.~abs/2111.06377,
  2021.

\bibitem{8953974}
J.~Fu, J.~Liu, H.~Tian, Y.~Li, Y.~Bao, Z.~Fang, and H.~Lu, ``Dual attention
  network for scene segmentation,'' in {\em 2019 IEEE/CVF Conference on
  Computer Vision and Pattern Recognition (CVPR)}, pp.~3141--3149, 2019.

\bibitem{SCHLEMPER2019197}
J.~Schlemper, O.~Oktay, M.~Schaap, M.~Heinrich, B.~Kainz, B.~Glocker, and
  D.~Rueckert, ``Attention gated networks: Learning to leverage salient regions
  in medical images,'' {\em Medical Image Analysis}, vol.~53, pp.~197--207,
  2019.

\bibitem{NEURIPS2021_64f1f27b}
E.~Xie, W.~Wang, Z.~Yu, A.~Anandkumar, J.~M. Alvarez, and P.~Luo, ``Segformer:
  Simple and efficient design for semantic segmentation with transformers,'' in
  {\em Advances in Neural Information Processing Systems} (M.~Ranzato,
  A.~Beygelzimer, Y.~Dauphin, P.~Liang, and J.~W. Vaughan, eds.), vol.~34,
  pp.~12077--12090, Curran Associates, Inc., 2021.

\bibitem{9711179}
W.~Wang, E.~Xie, X.~Li, D.-P. Fan, K.~Song, D.~Liang, T.~Lu, P.~Luo, and
  L.~Shao, ``Pyramid vision transformer: A versatile backbone for dense
  prediction without convolutions,'' in {\em 2021 IEEE/CVF International
  Conference on Computer Vision (ICCV)}, pp.~548--558, 2021.

\bibitem{8954457}
A.~Gupta, P.~Dollár, and R.~Girshick, ``Lvis: A dataset for large vocabulary
  instance segmentation,'' in {\em 2019 IEEE/CVF Conference on Computer Vision
  and Pattern Recognition (CVPR)}, pp.~5351--5359, 2019.

\bibitem{boykov2001interactive}
Y.~Y. Boykov and M.-P. Jolly, ``Interactive graph cuts for optimal boundary \&
  region segmentation of objects in nd images,'' in {\em Proceedings eighth
  IEEE international conference on computer vision. ICCV 2001}, vol.~1,
  pp.~105--112, IEEE, 2001.

\end{thebibliography}

\end{document}